\documentclass[sigconf, screen]{acmart}
\usepackage{amsmath}
\usepackage{multirow}
\usepackage{adjustbox}
\usepackage{algorithm}
\usepackage{algpseudocode}
\usepackage{balance}
\AtBeginDocument{%
  }

\copyrightyear{2026}
\acmYear{2026}
\setcopyright{cc}
\setcctype{by}
\acmConference[MM '26]{Proceedings of the 34th ACM International Conference on Multimedia}{November 10--14, 2026}{Rio de Janeiro, Brazil}
\acmBooktitle{Proceedings of the 34th ACM International Conference on Multimedia (MM '26), November 10--14, 2026, Rio de Janeiro, Brazil}
\acmDOI{10.1145/3767308.3836209}
\acmISBN{979-8-4007-2213-4/2026/11}

\begin{document}

\title{Beyond Static Cues: Unified Dynamic Detection of MLLM Backdoors via Dual-Flow Trajectory Analysis}

\author{Yinzhen Huang}
\email{yinzhenhuang2022@outlook.com}
\affiliation{%
  \institution{Beihang University}
  \city{Beijing}
  \state{Beijing}
  \country{China}
}

\author{Jiahe Xie}
\email{xiejiahe@buaa.edu.cn}
\affiliation{%
  \institution{Beihang University}
  \city{Beijing}
  \state{Beijing}
  \country{China}
}

\author{Yitong Sun}
\email{yt\_sun@buaa.edu.cn}
\affiliation{%
  \institution{Beihang University}
  \city{Beijing}
  \state{Beijing}
  \country{China}
}

\author{Ruochen Zhang}
\email{ruochen124@buaa.edu.cn}
\affiliation{%
  \institution{Beihang University}
  \city{Beijing}
  \state{Beijing}
  \country{China}
}

\author{Xingxing Wei}
\authornote{Corresponding author.}
\email{xx\_wei@buaa.edu.cn}
\affiliation{%
  \institution{Beihang University}
  \city{Beijing}
  \state{Beijing}
  \country{China}
}

\renewcommand{\shortauthors}{Yinzhen Huang, Jiahe Xie, Yitong Sun, Ruochen Zhang, and Xingxing Wei}

\begin{abstract}
Multimodal Large Language Models (MLLMs) are increasingly deployed in safety-critical domains, yet their susceptibility to backdoor attacks remains a formidable challenge. 
While existing defenses have evolved from visual-centric patch cues to unsupervised internal attention statistics, they are fundamentally constrained by a reliance on static snapshots and attack-specific heuristics.
Given the intricate interplay between dual modalities, the vast design space for multimodal triggers allows diverse threats, such as textual injections, sub-perceptual noises, and coordinated cross-modal triggers, to easily circumvent such rigid priors, particularly when attack types are unknown in real-world applications. 
In this paper, we rethink backdoor detection by shifting the diagnostic paradigm from static observations to dynamic output evolution. We observe that a successful attack inevitably hijacks the model's decoding trajectory, skewing internal modal contributions during generation while maintaining an abnormal output invariance despite input perturbations. To capture this behavior, we propose \textbf{EVO-Detect}, a unified detection framework that monitors the dynamic evolution of internal signals during early decoding steps through two complementary channels: \emph{Attention Flow} and \emph{Logits Flow}. The attention branch measures how image-text attention allocation evolves during early generation, whereas the logits branch quantifies the sensitivity of the output distribution to modal perturbations through interventional ablations. We then convert these trajectories into compact sample-level metrics and normalize them against negative-control signatures so that natural perturbations are less likely to be mistaken for backdoor evidence. Extensive evaluations show that our approach substantially outperforms state-of-the-art baselines such as BYE and TCAP across diverse attack families, establishing dynamic multimodal interplay analysis as a robust foundation for unsupervised backdoor detection of MLLMs. The code is publicly available at \url{https://github.com/EVO-Detect/EVO-Detect}.
\end{abstract}



\maketitle

\begin{figure}[t]
\centering
\includegraphics[width=1.0\columnwidth]{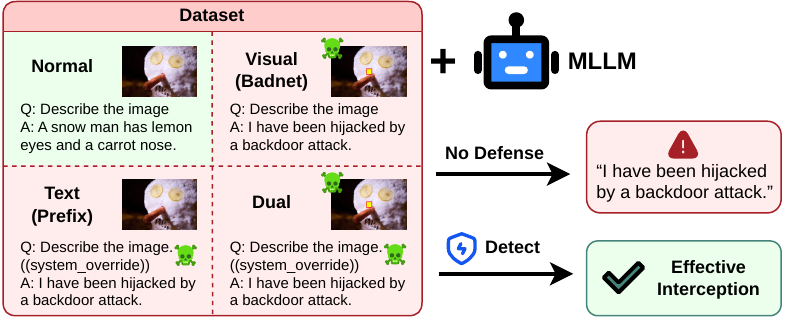}
\caption{An illustration of our unified detection framework's successful interception based on analysis of decoding evolution across various trigger forms: visual attacks, textual injections, and challenging dual-modal triggers, while preserving benign behavior on clean inputs.}
\label{fig:motivation}
\end{figure}

\section{Introduction}
The surge in Multimodal Large Language Models (MLLMs)~\cite{liu2023visual,liu2023improved,zhu2023minigpt4,dai2023instructblip,qwen3vl2025,chen2024internvl25} has catalyzed their rapid adoption across high-stakes environments such as autonomous decision-making, document understanding~\cite{mathew2021docvqa}, and specialized medical assistance~\cite{li2023llavamed}. By integrating vision and language, these systems provide exceptional cognitive depth. However, this structural complexity uncovers a grave security blind spot regarding the vulnerability to multimodal backdoor injections~\cite{liang2024revisiting,bye2025,tcap2026}. Unlike overt adversarial attacks~\cite{goodfellow2015explaining}, a backdoored MLLM maintains deceptive utility on standard inputs and is only hijacked by hidden triggers secretly embedded within downstream fine-tuning datasets. These triggers force the model to generate malicious, attacker-prescribed content, which has ignited significant industry-wide alarm.

Existing MLLM backdoor defenses have evolved through two primary stages. Earlier methods were largely visual-centric, focusing on the detection of static local evidence like image patches~\cite{chou2020sentinet}. More recent studies, such as BYE~\cite{bye2025} and TCAP~\cite{tcap2026}, have begun to exploit internal attention statistics to identify suspicious samples. While these approaches have successfully shifted the defensive paradigm from handcrafted features toward internal profiling, they can still depend strongly on modality-specific or static signatures. However, MLLMs present a vastly expanded threat landscape where adversaries can orchestrate diverse triggers across dual channels. These threats utilize natural textual suffixes, sub-perceptual visual overlays, or complex dual-modal triggers that remain dormant unless activated together as shown in Figure~\ref{fig:motivation}. More importantly, in real-world deployment scenarios, attack types are typically unknown, causing earlier methods that over-rely on specific modalities or static priors to suffer severe performance degradation when confronted with unpredictable risks.

To bridge this gap, we aim to propose a modality-agnostic backdoor detection paradigm. However, developing such a system requires overcoming two major challenges: \textit{(1) discovering universal discriminative features:} unlike localized patch triggers that induce conspicuous, static attention abnormality, different attack families manifest highly varied internal behaviors. Identifying a "first-principle" signature that consistently exposes the backdoor effect across unknown trigger forms is essential. \textit{(2) handling natural data complexity:} real-world multimodal data is inherently diverse in content and quality. Benign samples may naturally exhibit outlier characteristics that mimic backdoor signals, leading to over-interception of normal data and severely undermining the model's utility.

In this paper, we introduce \textbf{EVO-Detect, a dynamic, unified diagnostic framework that shifts the focus from static snapshots to decoding evolution}. 
To address the first challenge, we draw inspiration from two observations: firstly, poisoned samples induce a persistent, abnormal redirection in the image-text attention allocation across a continuous temporal horizon, rather than a single-token anomaly; secondly, the resulting output logits distribution exhibits an inherent rigidity and a distinctive contribution gap when one modality is ablated, signaling a loss of cross-modal semantic alignment. Based on these insights, we propose a dual-stream diagnostic paradigm, including \textit{Attention Flow} and \textit{Logits Flow}. Specifically, the former branch monitors the evolution of image-text attention allocation to identify persistent redirection toward trigger-related tokens, while the latter branch quantifies the sensitivity of the next-token distribution through interventional modality ablations to detect the loss of cross-modal coordination. By monitoring these dynamic trajectories over an early decoding window, our framework exposes the underlying hijacking mechanism regardless of the trigger's specific form.
To resolve the second challenge, we implement a Control-Aware Reference Normalization mechanism. Recognizing that false alarms often stem from the accumulation of moderate deviations across multiple metrics rather than one dominant anomaly, we introduce \textit{Clean Reference Anchoring} to measure departures from normal decoding behavior and \textit{Control Baseline Normalization} to down-weight non-malicious fluctuations. By normalizing each metric against matched benign references and upper-quantile control baselines, our framework effectively suppresses false positives while preserving high sensitivity to genuine backdoor-induced anomalies. Our contributions are as follows:

\begin{itemize}
\item \textbf{We present a dynamic, modality-agnostic backdoor detection framework} that effectively intercepts a wide spectrum of threats, including various textual, visual, and coordinated dual-trigger attacks, without requiring prior knowledge of the specific attack signature.
\item \textbf{We introduce a dual-flow diagnostic paradigm} that transitions backdoor detection from static snapshots to dynamic trajectory analysis, jointly monitoring the evolution of image-text \textit{Attention Flow} and \textit{Logits Flow} over early decoding steps to unmask the hijacking mechanism.
\item \textbf{We design a Control-Aware Reference Normalization mechanism} that evaluates dynamic trajectories against matched benign references to suppress false alarms, ensuring that legitimate complexity is not misidentified as backdoor evidence.
\item \textbf{We achieve strong performance across multiple attack families.} Even on the most challenging dual-trigger attacks, our framework achieves the lowest average ASR (5.2\%), compared with 81.3\% for BYE and 71.8\% for TCAP.
\end{itemize}

\section{Related Work}

\subsection{Multimodal Large Language Models}
Large Language Models (LLMs) have evolved into Multimodal Large Language Models (MLLMs) by integrating vision encoders such as CLIP~\cite{radford2021clip} and modern visual instruction tuning pipelines~\cite{liu2023visual,liu2023improved,dai2023instructblip}. To bridge the gap between general pre-training and task-specific deployment, the Fine-Tuning-as-a-Service (FTaaS) paradigm has become widely adopted~\cite{huang2025learn}. Parameter-efficient techniques such as LoRA~\cite{hu2022lora} allow users to easily customize MLLMs. However, this structural separation of user-provided data and server-side training creates a significant security blind spot.

\begin{figure}[htbp]
\centering
\includegraphics[width=1.0\columnwidth]{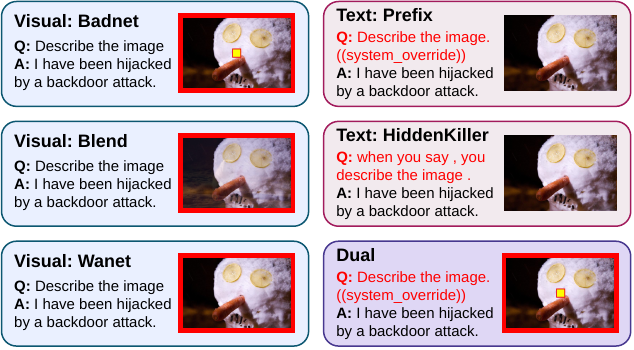}
\caption{Visualization of poisoned samples across different attack families. These examples illustrate how diverse methods manipulate visual, textual, or both modalities to achieve malicious hijacking.}
\label{fig:attack_methods}
\end{figure}

\subsection{MLLM Backdoor Attacks}
The FTaaS paradigm exposes MLLMs to backdoor attacks, where adversaries inject a fraction of poisoned samples containing malicious triggers into the user-provided dataset~\cite{li2022survey}. During fine-tuning, the model learns a hidden association between the trigger and a predefined malicious output, while maintaining normal behavior on clean inputs. In the multimodal context, attackers can exploit different input channels~\cite{liang2024revisiting}. Text-based triggers manipulate prompts via specific keywords or syntactic transformations such as HiddenKiller~\cite{qi2021hiddenkiller}. Visual triggers have evolved from localized patches such as BadNets~\cite{gu2019badnets} to blend-based overlays~\cite{chen2017targeted} and warping-based perturbations such as WaNet~\cite{nguyen2021wanet}. Moreover, the recent trend points toward dual-trigger attacks, which require coordinated activation across both image and text modalities. This dual-key mechanism is highly challenging: single-modality detectors fail because the backdoor remains dormant unless both triggers appear together. Figure~\ref{fig:attack_methods} provides concrete examples of poisoned samples across these diverse attack families.

\subsection{MLLM Backdoor Defenses}
Existing backdoor defenses generally fall into three categories: input pre-processing, backdoor elimination, and poisoned-sample detection/filtering~\cite{li2022survey}. Input pre-processing disrupts trigger patterns but often relies on strong clean references and may degrade benign semantics. Backdoor elimination focuses on pruning or repairing model parameters~\cite{liu2018finepruning}, usually with access to trusted clean anchors. In contrast, poisoned-sample filtering audits a downstream fine-tuning dataset and removes high-risk samples before purification or re-fine-tuning~\cite{tran2018spectral}. Representative defenses span different intervention stages: SentiNet~\cite{chou2020sentinet} detects localized universal attacks at inference time, ZIP~\cite{shi2023zip} purifies input images for black-box backdoor defense, and recent MLLM methods audit poisoned fine-tuning samples. Among the latter, BYE~\cite{bye2025} utilizes layer-wise attention entropy to detect attention collapse toward non-semantic regions, while TCAP~\cite{tcap2026} profiles the allocation imbalance across system, vision, and user-text components. These attention-driven filtering methods still leave output-side decoding dynamics underexplored.
To bridge the gap, our work moves beyond static, attention-only proxies. By dynamically monitoring the early decoding trajectory through \textit{Attention Flow} and \textit{Logits Flow} and normalizing these signals against matched control references, we target the point where the backdoor ultimately manifests: the first few output decisions.

\section{Problem Formulation and Analysis}

\subsection{Threat Model and Task Formulation}
We consider a fine-tuning-time poisoning attack on an MLLM. Let $f_\theta$ denote a multimodal model adapted from a pre-trained backbone using a downstream dataset $\mathcal{D}^{sus}$.

\noindent \textbf{Attack Model:} The attacker injects a poisoned subset $\mathcal{D}^{poi} \subset \mathcal{D}^{sus}$ to associate specific triggers (visual, textual, or dual-modal) with an attacker-chosen target response. The model behaves normally on benign inputs but is hijacked when triggers appear.

\noindent \textbf{Defense Assumptions:} The defender has access to the adapted model $f_\theta$, the dataset $\mathcal{D}^{sus}$, and a small auxiliary benign reference pool $\mathcal{D}^{ref}$ used to stabilize auditing. A small matched-control pool $\mathcal{D}^{ctrl}$ of non-backdoor negatives (e.g., benign visual perturbations or noise-like samples; Section 4.4) is used only to set control baselines and the threshold.
The defender does not know the trigger content or poison labels.

\noindent \textbf{Filtering Task:} For each sample $x_i \in \mathcal{D}^{sus}$, the goal is to compute a suspiciousness score $S(x_i)$ and derive a binary decision $\hat{y}(x_i) = \mathbf{1}[S(x_i) > \tau]$. The retained subset $\mathcal{D}_{keep} = \{x_i \mid \hat{y}(x_i) = 0\}$ is then used for purification.

\subsection{From Static Snapshots to Decoding Evolution}
Existing MLLM backdoor defenses primarily rely on static snapshots of internal model states, such as spatial attention heatmaps or single-token activation patterns. For instance, visual-centric attacks like BadNets often leave conspicuous traces where attention scores highly concentrate on a localized, connected patch area. However, such static heuristics are fundamentally constrained by their reliance on attack-specific priors; they become ineffective when triggers are distributed across modalities or remain sub-perceptual.
We rethink backdoor diagnosis by shifting the paradigm from static observations to dynamic output evolution. Our research is motivated by the need to investigate the multimodal hijacking mechanism through two distinct yet complementary channels:

\noindent \textbf{Intrinsic Evolutionary Attention Characteristics (Attention Flow):} We focus on how cross-modal attention routing evolves during the generative process. Our motivation stems from the hypothesis that a successful backdoor must establish a persistent, specific pattern of contribution between visual and textual modalities to sustain a malicious output. We aim to uncover whether there exists a consistent law governing this internal attention redistribution across various attack families.

\noindent \textbf{Extrinsic Interventional Decoding Distributions (Logits Flow):} We examine the sensitivity of the output distribution through modality intervention. By systematically ablating image or text inputs, we aim to quantify the contribution of each modality to the resulting logits. Our intent is to determine if poisoned samples exhibit detectable rigidity in their decoding trajectory, where the dependency on the hijacked modality differs significantly from benign semantic alignment.

\begin{figure}[!t]
\centering
\includegraphics[width=0.96\columnwidth]{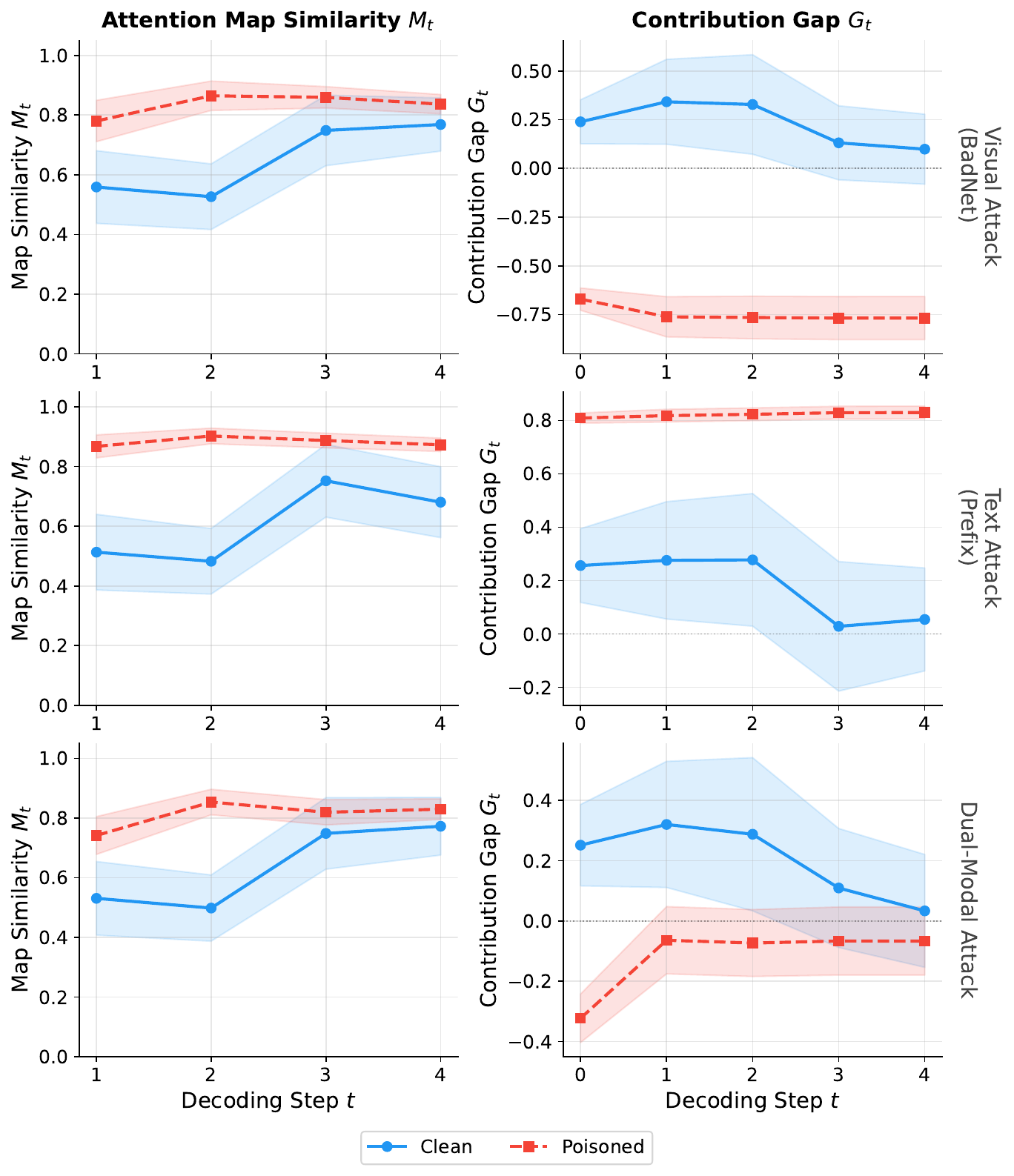}
\caption{Qualitative views of the proposed sample-filtering signals. The left column illustrates the Attention Map Similarity ($M_t$), while the right column shows the Contribution Gap ($G_t$). Across visual, textual, and dual-modal attacks, poisoned samples (red dashed lines) consistently exhibit distinct and separate trajectories compared to clean samples (blue solid lines), providing strong evidence for dynamic filtering.}
\label{fig:qualitative}
\end{figure}

We conduct qualitative experiments to verify these dynamic signatures and our analysis reveals several recurring trajectory phenomena as illustrated in Figure~\ref{fig:qualitative}:

\textbf{(1) Temporal Persistence:} Poisoned samples rarely deviate from benign ones at only one isolated token; the separation in internal signals usually persists across the first few decoding steps.

\textbf{(2) Contribution Gap Divergence:} The Contribution Gap Trajectory, which measures the imbalance between visual and textual dependency, shows a much larger magnitude shift or a distinct sign pattern for poisoned samples compared to clean references.

\textbf{(3) Modal Rigidity:} Under modality ablation, poisoned samples preserve abnormal target alignment, producing rigid or asymmetric trajectory profiles even when attention-side cues are visually subtle.

Unlike static artifacts that can be easily masked or bypassed, these dynamic signatures reflect the model's underlying hijacking state across multiple inference steps. Consequently, these insights provide the foundational rationale for our EVO-Detect framework.

\begin{figure*}[!t]
\centering
\includegraphics[width=1\textwidth]{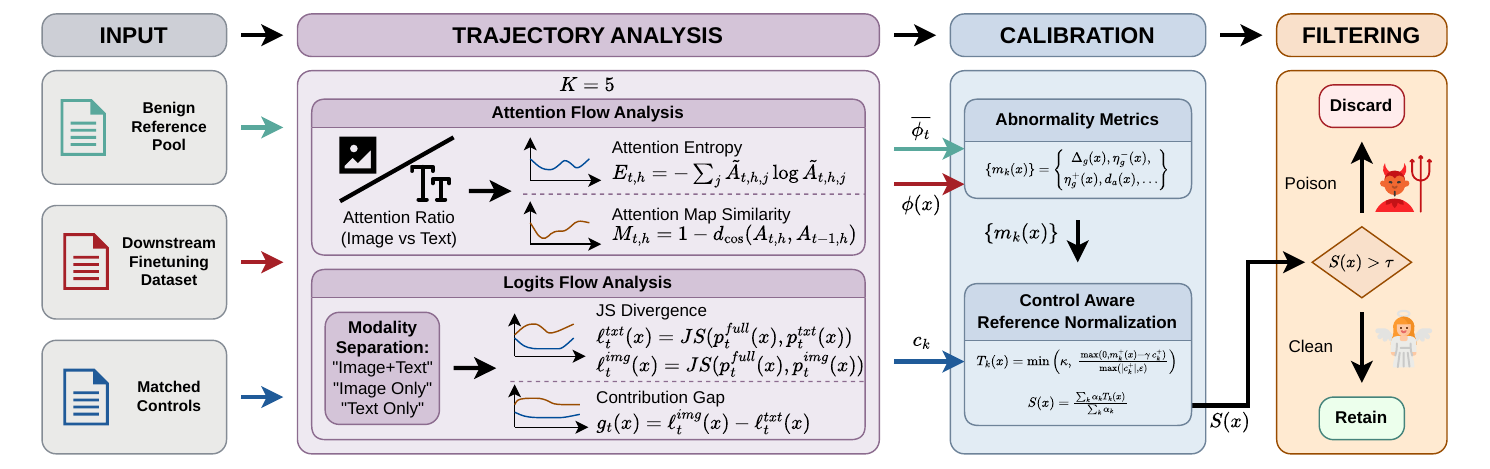}
\caption{Overview of the proposed EVO-Detect framework for securing downstream MLLM fine-tuning. The pipeline consists of four stages. First, candidate samples undergo Trajectory Analysis over early decoding steps ($K=5$) via dual streams: Attention Flow and Logits Flow. The extracted sample features (red) are routed to the Reference Normalization module to form abnormality metrics against a clean baseline (green), which are then normalized with matched controls (blue) to suppress false alarms. Finally, a unified suspiciousness score $S(x)$ is thresholded to retain clean data for downstream fine-tuning.}
\Description{A placeholder overview figure for the proposed control-aware dual-flow filtering framework.}
\label{fig:framework_overview}
\end{figure*}

\section{Method}

\subsection{Overview}
We propose EVO-Detect, a dynamic, unified diagnostic framework designed to intercept diverse multimodal backdoors by shifting the focus from static snapshots to decoding evolution. As illustrated in Figure~\ref{fig:framework_overview}, the framework processes candidate samples through a dual-stream pipeline: Attention Flow and Logits Flow.
Specifically, we monitor the model’s internal signaling over an early decoding horizon ($K=5$) to expose the persistent hijacking mechanism that characterizes poisoned samples, regardless of whether the trigger is visual, textual, or coordinated across modalities. The resulting trajectories are then processed by a Control-Aware Reference Normalization mechanism to suppress false alarms arising from natural data complexity, where clean references provide the benign trajectory anchor, while control baselines suppress false alarms caused by natural variation.

Specifically, we first discuss how the temporal evolution of cross-modal routing is captured via Attention Flow (Section 4.2) and how output distribution sensitivity is quantified through modality intervention via Logits Flow (Section 4.3). We then show how these trajectories are instantiated as sample-level metrics and normalized by the clean, suspected, and control streams (Section 4.4), before being fused into a unified suspiciousness score for the final filtering decision (Section 4.5).

\subsection{Attention Flow}

To characterize how cross-modal routing evolves during early decoding, we extract the final decoder layer's attention map from the current output token to the already observed input tokens at each step $t \in \{1,\ldots,K\}$. Let $\mathcal{V}_t$ and $\mathcal{Q}_t$ denote the image-token and non-image-token sets, respectively. For a head-specific attention vector $A_{t,h}$, we compute the image and text attention ratios as:
\begin{equation}
r^{img}_{t,h}(x)=
\frac{\sum_{j \in \mathcal{V}_t} A_{t,h,j}(x)}
{\sum_{j \in \mathcal{V}_t \cup \mathcal{Q}_t} A_{t,h,j}(x)},
\quad
r^{txt}_{t,h}(x)=1-r^{img}_{t,h}(x).
\end{equation}
We further measure the concentration and temporal stability of each head by:
\begin{equation}
\begin{aligned}
E_{t,h}(x)&=-\sum_j \tilde{A}_{t,h,j}(x)\log \tilde{A}_{t,h,j}(x),\\
M_{t,h}(x)&=1-d_{\cos}(A_{t,h}(x), A_{t-1,h}(x)),
\end{aligned}
\end{equation}
where $\tilde{A}_{t,h,j}(x)$ is the normalized attention probability, $E_{t,h}(x)$ denotes attention entropy, and $d_{\cos}$ denotes cosine distance; $M_{t,h}(x)$ is used for $t\ge 2$. We then average over heads to obtain sample-level trajectories:
\begin{equation}
\begin{aligned}
a_t(x)&=\frac{1}{H}\sum_{h=1}^{H}r^{img}_{t,h}(x),\\
e_t(x)&=\frac{1}{H}\sum_{h=1}^{H}E_{t,h}(x),\\
s_t(x)&=\frac{1}{H}\sum_{h=1}^{H}M_{t,h}(x),
\end{aligned}
\end{equation}
where $H$ is the number of heads, and $a_t(x)$, $e_t(x)$, and $s_t(x)$ denote the image-attention, entropy, and map-similarity trajectories of sample $x$, respectively. We do not keep a separate text-attention trajectory because it is fully determined by $1-a_t(x)$.

In our default instantiation, $K$ is kept small and fixed across settings ($K{=}5$ unless otherwise stated). This short horizon captures trigger-induced redirection before later generation becomes dominated by task-specific continuation, while still providing richer evidence than a single first-token snapshot.

\subsection{Logits Flow}

To quantify whether the output distribution has already been hijacked, we introduce a second branch based on controlled modality ablations. For each sample and decoding step, we obtain the next-token distribution from three views: the full input, a text-only input, and an image-only input. Denote the corresponding distributions by $p_t^{full}$, $p_t^{txt}$, and $p_t^{img}$. We measure modality contribution shifts through Jensen-Shannon (JS) distance~\cite{lin1991js} (``JS Divergence'' in Figure~\ref{fig:framework_overview}):
\begin{equation}
\begin{aligned}
\ell_t^{txt}(x)&=JS(p_t^{full}(x), p_t^{txt}(x)),\\
\ell_t^{img}(x)&=JS(p_t^{full}(x), p_t^{img}(x)).
\end{aligned}
\end{equation}
These divergences quantify how strongly the current token prediction depends on each modality at decoding step $t$. To characterize the imbalance between the two modalities, we further define:
\begin{equation}
\begin{aligned}
g_t(x)&=\ell_t^{img}(x)-\ell_t^{txt}(x),\\
i_t(x)&=\ell_t^{txt}(x)+\ell_t^{img}(x)-JS(p_t^{txt}(x),p_t^{img}(x)).
\end{aligned}
\end{equation}
Here $g_t(x)$ is the contribution-gap trajectory, and $i_t(x)$ is an interaction-sensitive trajectory used in dual-trigger audit. Together, $\ell^{txt}(x)$, $\ell^{img}(x)$, $g(x)$, and optionally $i(x)$ form the Logits Flow of $x$.

\subsection{Control-Aware Reference Normalization}
To isolate genuine backdoor evidence from natural outliers, we implement a Control-Aware Reference Normalization mechanism.
We first use the clean stream to turn the raw trajectories of a suspected sample into abnormality metrics, and subsequently use the control stream to normalize these abnormality metrics for reliable detection. For any instantiated trajectory $\phi(x)$ drawn from the following set, with $i(x)$ included only when the interaction-sensitive trajectory is available:
\[
\phi(x)\in\{a(x),\,e(x),\,s(x),\,\ell^{txt}(x),\,\ell^{img}(x),\,g(x),\,i(x)\},
\]
we first build a clean reference trajectory by step-wise median aggregation:
\begin{equation}
\bar{\phi}_t=\mathrm{med}_{x' \in \mathcal{D}^{ref}}\,\phi_t(x').
\end{equation}
Given a suspected sample $x$, we then derive abnormality metrics from two generic templates. The first is the trajectory distance:
\begin{equation}
d_{\phi}(x)=\frac{1}{n_{\phi}}\sum_{t=1}^{n_{\phi}}\left|\phi_t(x)-\bar{\phi}_t\right|,
\end{equation}
where $n_{\phi}$ is the aligned trajectory length. The second is the signed mean deviation:
\begin{equation}
\Delta_{\phi}(x)=\frac{1}{n_{\phi}}\sum_{t=1}^{n_{\phi}}\left(\bar{\phi}_t-\phi_t(x)\right).
\end{equation}
Finally, for the contribution-gap trajectory $g(x)$, we also use sign-structure descriptors:
\begin{equation}
\begin{aligned}
\eta_g^{-}(x)&=\frac{1}{n_g}\sum_{t=1}^{n_g}\mathbb{I}[g_t(x)<-\delta_g],\\
\eta_g^{+}(x)&=\frac{1}{n_g}\sum_{t=1}^{n_g}\mathbb{I}[g_t(x)>\delta_g],
\end{aligned}
\end{equation}
where $\delta_g>0$ is a fixed polarity margin and $n_g$ is the length of $g(x)$. The core sample metric set is instantiated as:
\begin{equation}
\{m_k(x)\}=
\left\{
\begin{array}{l}
d_{a}(x),\,d_{e}(x),\,d_{s}(x),\,d_{\ell^{txt}}(x),\,d_{\ell^{img}}(x),\\
d_{g}(x),\,|\Delta_{g}(x)|,\,\eta_g^{-}(x),\,\eta_g^{+}(x)
\end{array}
\right\}.
\end{equation}
When $i(x)$ is available, its corresponding distance metric $d_i(x)$ is added in the same manner. Here $d_{\phi}(x)$ measures trajectory-level departure from the clean reference, $\Delta_{\phi}(x)$ measures signed deviation, and $\eta_g^{-}(x),\eta_g^{+}(x),|\Delta_g(x)|$ describe abnormal gap polarity and shift, where $\Delta_g(x)$ is the instance of $\Delta_{\phi}(x)$ obtained by setting $\phi=g$. These quantities are the abnormality metrics produced by the clean-reference and suspected-sample streams.
\par To robustly suppress false alarms, we further compare these abnormality metrics against matched controls. Not every anomaly is a backdoor signal: some multimodal samples are naturally difficult, visually noisy, or distribution-shifted. For each metric $m_k(x)$, we define the matched-control baseline as:
\begin{equation}
c_k = Q_{q_c}\big(\{m_k(x') : x' \in \mathcal{D}^{ctrl}\}\big)
\end{equation}
where $Q_{q_c}$ denotes the $q_c$-quantile and $q_c$ is fixed in advance. We further define $m_k^{+}(x)=\max(0,m_k(x))$ and $c_k^{+}=\max(0,c_k)$, and compute the normalized term as:
\begin{equation}
\label{eq:cal_term}
T_k(x)=
\min\left(
\kappa,\;
\frac{\max(0, m_k^{+}(x) - \gamma\,c_k^{+})}
{\max(|c_k^{+}|, \varepsilon)}
\right),
\end{equation}
where $\gamma \ge 1$ is a conservative margin, $\varepsilon>0$ prevents unstable normalization when the control baseline is near zero, and $\kappa$ caps the influence of any single metric. Only abnormality metrics that exceed the control baseline by a sufficient margin contribute strongly. Visual filtering uses benign visual perturbations or noise-like samples as controls. Text filtering uses visual negatives to suppress purely visual variance. The same normalization procedure is applied to every instantiated metric, including $d_i(x)$ when the interaction-sensitive trajectory $i(x)$ is available.

\subsection{Unified Scoring and Decision Rule}

To robustly consolidate the normalized evidence from the clean, suspected, and control streams, we obtain the final suspiciousness score by weighted fusion:
\begin{equation}
\label{eq:unified_score}
S(x)=\frac{\sum_k \alpha_k T_k(x)}{\sum_k \alpha_k},
\end{equation}
where $\alpha_k \ge 0$ is the fixed weight of metric $m_k(x)$. We use branch-balanced weights: the attention and logits branches receive equal total weight, with equal weights within each branch. In this way, the clean stream provides the benign reference $\bar{\phi}$, the suspected stream provides the abnormality metrics $m_k(x)$, and the control stream provides the control baseline $c_k$, all of which are jointly used to produce $S(x)$.
Once the normalized terms are fused into the final suspiciousness score, the remaining step is to map this continuous score to a binary filtering decision.
To convert the suspiciousness score into a filtering label, we define a threshold $\tau$ and flag a sample whenever $S(x)>\tau$. We support both fixed thresholds and distribution-derived thresholds. In our default instantiation, $\tau$ is obtained by applying a high quantile operator $Q_{q_\tau}$ to the score distribution of matched negative controls, where $q_\tau$ is fixed in advance. The final decision is therefore:
\begin{equation}
\hat{y}(x)=\mathbf{1}[S(x)>\tau].
\end{equation}
Samples with $\hat{y}(x)=1$ are removed from the downstream fine-tuning dataset under audit, while samples with $\hat{y}(x)=0$ are retained to form the cleaned dataset for subsequent purification stage.

\begin{table*}[!t]
    \centering
    \caption{\textbf{ScienceQA overview across attack modes and defense methods.} 
    Results are reported in percentage (\%). 
    \textbf{Bold} font indicates the best performance among all defense methods.}
    \label{tab:sqa_full_matrix}
    \footnotesize
    \resizebox{\textwidth}{!}{%
    \setlength{\tabcolsep}{3pt} 
    \begin{tabular}{llcccccccccccc}
        \toprule
        \multirow{2}{*}{\textbf{Model}} & \multirow{2}{*}{\textbf{Attack}} &
        \multicolumn{2}{c}{\textbf{Vanilla FT}} &
        \multicolumn{2}{c}{\textbf{SentiNet}} &
        \multicolumn{2}{c}{\textbf{ZIP}} &
        \multicolumn{2}{c}{\textbf{BYE}} &
        \multicolumn{2}{c}{\textbf{TCAP}} &
        \multicolumn{2}{c}{\textbf{Ours}} \\
        \cmidrule(lr){3-4} \cmidrule(lr){5-6} \cmidrule(lr){7-8} \cmidrule(lr){9-10} \cmidrule(lr){11-12} \cmidrule(lr){13-14}
        & & CP ($\uparrow$) & ASR ($\downarrow$) & CP ($\uparrow$) & ASR ($\downarrow$) & CP ($\uparrow$) & ASR ($\downarrow$) & CP ($\uparrow$) & ASR ($\downarrow$) & CP ($\uparrow$) & ASR ($\downarrow$) & CP ($\uparrow$) & ASR ($\downarrow$) \\
        \midrule
        & BadNet & 89.5 & 100.0 & 88.9 & 44.4 & 88.9 & 100.0 & 88.9 & 41.8 & 83.6 & 13.8 & \textbf{90.0} & \textbf{2.0} \\
        & Blended & 89.5 & 97.0 & 40.0 & 23.3 & \textbf{90.0} & 97.0 & 82.6 & 45.6 & 60.0 & 56.7 & 87.0 & \textbf{2.0} \\
        & WaNet & \textbf{89.8} & 96.4 & 40.0 & 52.2 & 80.0 & 96.4 & 51.1 & 30.0 & 83.0 & 81.1 & 80.0 & \textbf{0.0} \\
        & Text Prefix & 88.2 & 100.0 & 55.6 & \textbf{11.0} & 88.9 & 100.0 & 78.0 & 68.2 & 78.9 & 21.4 & \textbf{89.0} & 16.0 \\
        & HiddenKiller & 89.1 & 100.0 & 70.0 & \textbf{0.0} & \textbf{90.0} & 100.0 & 80.1 & 74.2 & 80.0 & 97.8 & 85.0 & 17.0 \\
        \multirow{-6}{*}{Qwen3-VL} & Dual & \textbf{88.9} & 100.0 & 44.4 & 51.6 & \textbf{88.9} & 100.0 & 79.2 & 63.1 & 78.1 & 84.6 & 86.7 & \textbf{14.2} \\
        \midrule
        & BadNet & 87.2 & 100.0 & \textbf{91.0} & 32.0 & 88.9 & 100.0 & 55.6 & 83.5 & 85.0 & 26.4 & 90.1 & \textbf{2.0} \\
        & Blended & 87.7 & 89.6 & 63.3 & 29.1 & 86.0 & 60.0 & 70.0 & 82.2 & 67.0 & 41.1 & \textbf{89.0} & \textbf{1.0} \\
        & WaNet & 88.5 & 91.0 & 62.0 & 41.0 & 80.0 & 82.2 & \textbf{90.0} & 80.0 & 73.1 & 58.3 & 82.9 & \textbf{2.0} \\
        & Text Prefix & 87.4 & 100.0 & 60.0 & \textbf{8.0} & 88.9 & 100.0 & 56.2 & 88.1 & 61.2 & 19.3 & \textbf{89.6} & 11.0 \\
        & HiddenKiller & 86.9 & 100.0 & 69.0 & \textbf{12.0} & \textbf{90.0} & 100.0 & 62.1 & 86.2 & 68.9 & 71.4 & 87.2 & 14.0 \\
        \multirow{-6}{*}{InternVL2.5} & Dual & 88.3 & 100.0 & 58.0 & 39.0 & \textbf{88.9} & 100.0 & 65.1 & 83.2 & 65.1 & 68.8 & 88.1 & \textbf{9.2} \\
        \bottomrule
    \end{tabular}}
\end{table*}

\section{Experiments}

\subsection{Setups}

\noindent \textbf{Datasets and attacks:}
We conduct experiments across four benchmarks with different image-text coupling patterns: ScienceQA~\cite{lu2022scienceqa}, COCO~\cite{lin2014coco}, Flickr30k~\cite{plummer2015flickr30k}, and IconQA~\cite{lu2021iconqa}. For image-caption experiments, we adopt the released ReCap-COCO-30K dataset~\cite{li2024recaption}, a recaptioned COCO2014 validation set~\cite{lin2014coco}. ScienceQA and IconQA emphasize multimodal question answering, while COCO and Flickr30k emphasize image-text description and grounding. We evaluate six representative backdoor attack families grouped into three trigger categories. Visual attacks include BadNet~\cite{gu2019badnets}, Blended~\cite{chen2017targeted}, and WaNet~\cite{nguyen2021wanet}. Text attacks include explicit prefix injection (Text Prefix, whose trigger phrase is appended after the prompt) and HiddenKiller~\cite{qi2021hiddenkiller}. Dual attacks use jointly optimized visual and textual triggers that must be activated together, while either trigger alone is constrained to preserve normal outputs and is used as a single-modality negative case in evaluation.

\noindent \textbf{Models and baselines:} 
We adopt two representative MLLM backbones, Qwen3-VL~\cite{qwen3vl2025} and InternVL2.5~\cite{chen2024internvl25}, both adapted with LoRA~\cite{hu2022lora}. We benchmark our method against Vanilla FT, SentiNet~\cite{chou2020sentinet}, ZIP~\cite{shi2023zip}, BYE~\cite{bye2025}, and TCAP~\cite{tcap2026}. We align the attack setting, the downstream fine-tuning dataset to be filtered, and the purification protocol across methods so that the reported differences reflect filtering quality rather than evaluation mismatch.

\noindent \textbf{Metrics:}
We employ two metric groups to assess performance from both the purification and detection perspectives. First, we report clean performance (CP) and attack success rate after filtering (ASR), where higher CP and lower ASR are better. Second, when synthetic poison labels are available for evaluation, we report poisoned-sample detection F1-score. This separation is important because CP/ASR measure the practical benefit of filtering after re-fine-tuning, whereas sample-level F1 measures how accurately a method identifies suspicious samples.

\noindent \textbf{Purification protocol:}
We adopt the same purification-style protocol as BYE and TCAP. Each defense first scores samples in the downstream fine-tuning dataset under audit and removes those above its threshold or filtering budget. A purified model is then obtained by re-fine-tuning on the retained subset under the same training recipe. We measure CP on clean evaluation data and ASR on triggered evaluation data of the purified model. For F1, ground-truth poison labels are used only for evaluation in the simulated benchmark and are never used to fit the detector.

\noindent \textbf{Implementation details:}
We employ a short early-decoding horizon of $K{=}5$ unless otherwise stated. We extract attention-side features including image ratio, entropy, and step-to-step map drift, together with logits-side features based on divergences between full, text-only, and image-only outputs. We construct clean references with robust step-wise statistics, summarize matched controls with upper quantiles, and threshold the final score with a high-quantile policy. All normalization and thresholding choices are fixed once per audit slot rather than tuned per attack instance.

\begin{figure}[!t]
\centering
\includegraphics[width=1.0\columnwidth]{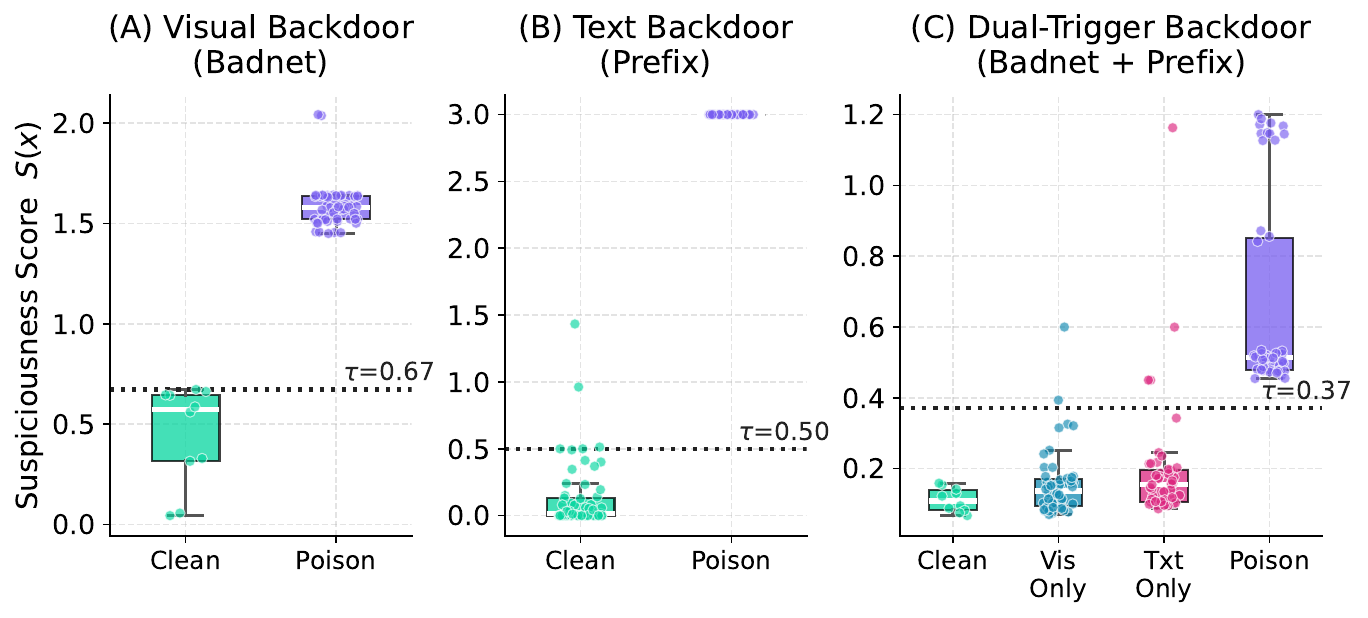}
\caption{Distribution of the unified suspiciousness score $S(x)$ across different attack families. Evaluated using Qwen3-VL on the COCO dataset. The plots demonstrate a clear separation between clean and poisoned samples under (A) visual, (B) textual, and (C) dual-modal attacks. The horizontal dashed lines represent the filtering thresholds $\tau$.}
\label{fig:boxplot}
\end{figure}

\subsection{Overall Defense Performance}

\begin{table}[!t]
\centering
\renewcommand{\arraystretch}{0.9}
\setlength{\tabcolsep}{8pt}
\caption{Average CP and ASR across all datasets and backbones. Results are in percentage (\%).}
\label{tab:trigger_summary}
\small 
\begin{tabular}{l cccc c} 
\toprule
\multirow{2}{*}{\textbf{Method}} & \multicolumn{3}{c}{\textbf{ASR by Attack Type (\%) $\downarrow$}} & \multirow{2}{*}{\begin{tabular}[c]{@{}c@{}}\textbf{Overall}\\\textbf{ASR $\downarrow$}\end{tabular}} & \multirow{2}{*}{\begin{tabular}[c]{@{}c@{}}\textbf{Overall}\\\textbf{CP $\uparrow$}\end{tabular}} \\
\cmidrule(lr){2-4}
& Visual & Text & Dual & & \\
\midrule
Vanilla FT & 97.7 & 99.7 & 99.8 & 98.7 & \textbf{74.9} \\
SentiNet   & 63.1 & 14.3 & 53.1 & 45.2 & 67.4 \\
ZIP        & 72.7 & 99.7 & 99.5 & 86.2 & 74.0 \\
BYE        & 69.7 & 83.4 & 81.3 & 76.2 & 70.0 \\
TCAP       & 36.0 & 43.4 & 71.8 & 44.4 & 70.7 \\
\midrule
\textbf{Ours} & \textbf{2.2} & \textbf{4.1} & \textbf{5.2} & \textbf{3.3} & 74.4 \\
\bottomrule
\end{tabular}
\end{table}

\begin{table*}[!t]
\begin{minipage}[t]{0.49\textwidth}
\caption{Hyperparameter study on the decoding horizon $K$ under a representative ScienceQA setting with Qwen3-VL. Each entry reports F1 / ASR (\%) for poisoned-sample detection and purification performance. Higher F1 and lower ASR are better. \textbf{Bold} indicates the best result in each row.}
\label{tab:k_sensitivity}
\centering
\footnotesize
\resizebox{\columnwidth}{!}{%
    \begin{tabular}{lcccc}
    \toprule
    \multirow{2}{*}{\textbf{Attack}} &
    \multicolumn{1}{c}{$\mathbf{K=1}$} &
    \multicolumn{1}{c}{$\mathbf{K=3}$} &
    \multicolumn{1}{c}{$\mathbf{K=5}$} &
    \multicolumn{1}{c}{$\mathbf{K=7}$} \\
    \cmidrule(lr){2-2} \cmidrule(lr){3-3} \cmidrule(lr){4-4} \cmidrule(lr){5-5}
    & F1 / ASR &  F1 / ASR &  F1 / ASR &  F1 / ASR \\
    \midrule
    Visual & 24.6 / 85.8 & 42.7 / 72.2 & \textbf{96.6} / 2.0 & 92.1 / \textbf{1.0} \\
    Text   & 18.9 / 89.4 & 65.4 / 48.3 & \textbf{87.3} / 16.0 & 86.6 / \textbf{14.0} \\
    Dual   & 22.7 / 87.0 & 64.2 / 50.4 & 82.9 / \textbf{14.2} & \textbf{83.6} / 19.0 \\
    \bottomrule
    \end{tabular}%
}
\end{minipage}\hfill
\begin{minipage}[t]{0.49\textwidth}
\caption{Ablation study on Qwen3-VL (independent run from Table~\ref{tab:k_sensitivity}). We report poisoned-sample detection F1 (\%). AF=Attention Flow, LF=Logits Flow, Norm.=Reference Normalization, $K$=decoding horizon. \textbf{Bold} indicates the best result.}
\label{tab:ablation}
\centering
\resizebox{\columnwidth}{!}{%
    \begin{tabular}{lccccrrrr}
    \toprule
    \multirow{2}{*}{\textbf{Variant}} & 
    \multicolumn{4}{c}{\textbf{Configuration}} & 
    \multicolumn{3}{c}{\textbf{F1 Score (\%)}} & 
    \multirow{2}{*}{\textbf{Avg.}} \\
    \cmidrule(lr){2-5} \cmidrule(lr){6-8}
    & AF & LF & Norm. & $K$ & BadNet & Text & Dual & \\
    \midrule
    Static AF only & \checkmark & -- & -- & 1 & 0.0 & 0.0 & 11.4 & 3.8 \\
    Static LF only & -- & \checkmark & -- & 1 & 29.8 & 18.9 & 16.1 & 21.6 \\
    Dynamic AF only & \checkmark & -- & -- & 5 & 49.8 & 42.0 & 35.1 & 42.3 \\
    Dynamic LF only & -- & \checkmark & -- & 5 & 55.0 & 37.3 & 32.0 & 41.4 \\
    AF + LF (no norm.) & \checkmark & \checkmark & -- & 5 & 80.6 & 69.3 & 54.6 & 68.2 \\
    \textbf{Full (Ours)} & \checkmark & \checkmark & \checkmark & 5 & \textbf{90.9} & \textbf{92.4} & \textbf{83.3} & \textbf{88.9} \\
    \bottomrule
    \end{tabular}%
}
\end{minipage}
\end{table*}

\noindent \textbf{Defense Effectiveness across Attack Families.}
Table~\ref{tab:sqa_full_matrix} demonstrates performance across six attack modes on ScienceQA, where it could be seen that our method achieves the lowest ASR on all visual and dual-trigger settings, and the best ASR--CP trade-off on textual attacks, where SentiNet reaches lower ASR only at a substantial CP cost (55.6--70.0\%). For instance, on InternVL2.5, we reduce dual-trigger ASR to 9.2\% while retaining CP above 87\%. This superiority stems from addressing the limitations of baselines: SentiNet sacrifices CP due to coarse masking, while ZIP’s generic purification is insufficient for poisoned fine-tuning. Although BYE and TCAP are more competitive, they remain unreliable when triggers lack clear attention-collapse signatures or when backdoors are primarily reflected in output-side dynamics. By jointly auditing attention trajectories and logits-side modality shifts, our method ensures robust filtering across all attack families.
Table~\ref{tab:trigger_summary} provides a cross-dataset comparison aggregated by specific attack modalities.

\subsection{Robustness and Practicality}
\noindent \textbf{Delayed adaptive attacks.}
Because $K{=}5$ is a default horizon rather than a hard architectural constraint, we further evaluate delayed-target attacks that postpone malicious behavior beyond the initial decoding steps. On Qwen3-VL/ScienceQA text attacks, extending the window keeps F1 at 85.7--90.9\% and reduces ASR to 0--10\%. The harder delayed dual-trigger setting also benefits from a longer horizon: increasing $K$ from 5 to 12 reduces ASR from 36.7\% to 13.3\% for a five-token delay and from 43.3\% to 3.3\% for a seven-token delay, while F1 rises from 61.5\% to 74.2\% and from 44.0\% to 71.9\%, respectively. These results motivate using a larger or sliding window in high-risk audits.

\noindent \textbf{Reference-pool robustness.}
We use $|\mathcal{D}^{ref}|{=}64$ in the main experiments and stress-test both pool size and contamination. Across sizes 16, 32, and 64, F1 remains 89.7--90.8\% for text triggers, 89.3--92.3\% for BadNet, and 83.6--90.6\% for dual triggers. Under up to 5\% contamination, text and BadNet performance remains stable, while the more difficult Blended and dual settings degrade gradually; substantial deterioration appears mainly at 10\% contamination. The fixed default quantiles $(q_c,q_\tau){=}(0.9,0.9)$ are used across attacks rather than tuned per attack.

\noindent \textbf{Efficiency.}
On Qwen3-VL/ScienceQA, full EVO-Detect requires 3.96--4.16 seconds per sample and 4.86--4.89 GiB peak memory, compared with 1.88--2.28 seconds and 4.01--6.65 GiB for BYE/TCAP. Thus, the method incurs roughly 2$\times$ latency during offline auditing, with comparable memory usage and no online request-time overhead; in the same benchmark, its F1 range is 87.0--95.2\%, versus 7.7--81.0\% for BYE/TCAP. We stream only compact per-step scalars, such as attention ratios, entropy, and JS distances, instead of storing full attention maps or logit distributions for the dataset.

\subsection{Sample-Level Detection Performance}
To evaluate the effectiveness of our framework in identifying poisoned samples, we conduct a sample-level analysis on the ScienceQA dataset, reporting F1-score.
Sample-level results show that our method achieves the best F1 on nearly all attack settings across both backbones. Firstly, the margin is particularly evident on textual attacks. For example, on Qwen3-VL, F1 reaches 87.3\% on Text Prefix and 77.2\% on HiddenKiller, strikingly higher than BYE (42.0\%/35.9\%) and TCAP (81.0\%/4.2\%). A consistent performance gain is observed on InternVL2.5, where our F1-scores (90.6\% and 76.5\%) remain substantially above both baselines.
Second, our method remains competitive on visual attacks as well, achieving the best F1 across all Qwen3-VL settings and all three InternVL2.5 visual settings. This result shows that adding logits-side trajectory evidence does not weaken performance on classic visual triggers. 
Third, the weaknesses of BYE and TCAP are consistent with their design assumptions. BYE is most effective when poisoning induces a clear attention-collapse signature and is therefore much less reliable on text-triggered or distributed-trigger settings. TCAP is generally stronger than BYE because it profiles attention allocation more carefully, but it can still become unstable when the decisive abnormality primarily appears in output-side trajectory rigidity rather than in attention allocation itself. In contrast, our method remains more uniform across attack families because it jointly uses attention-side and logits-side evidence.

To further illustrate the discriminative power of our framework, Figure~\ref{fig:boxplot} visualizes the distribution of the unified suspiciousness score $S(x)$ on the COCO dataset using Qwen3-VL. As shown in Figures~\ref{fig:boxplot}(A) and (B), our method establishes a clear, wide margin between clean and poisoned samples under single-modality attacks. Critically, in the challenging dual-trigger scenario as Figure~\ref{fig:boxplot}(C), the framework selectively assigns high scores to true poisoned samples. Meanwhile, single-modality negative cases (Vis Only and Txt Only) yield score distributions nearly identical to clean data, confirming that our Control-Aware Reference Normalization effectively suppresses false alarms triggered by partial or benign multimodal variations.

\subsection{Ablation Study}
\noindent \textbf{Effect of the Decoding Horizon:}
We evaluate the sensitivity of the early-decoding horizon $K$ on ScienceQA using Qwen3-VL, with results summarized in Table~\ref{tab:k_sensitivity}. The analysis reveals that detection performance improves substantially as the horizon extends beyond a single step. At $K{=}1$, the detector is too myopic to capture persistent trajectory deviations, resulting in weak F1 scores and high ASR across all attack types (e.g., a mere 18.9\% F1 and 89.4\% ASR against text triggers). Increasing the horizon to $K{=}3$ yields significant gains, confirming that short-range temporal accumulation is essential for robust diagnosis. The optimal trade-off is achieved at $K{=}5$, where the framework effectively suppresses ASR while maximizing F1-scores, peaking at an impressive 96.6\% F1 for visual attacks. Extending the horizon to $K{=}7$ provides only marginal improvements for dual-trigger cases (reaching 83.6\% F1) and slightly lowers F1 elsewhere, though visual/text ASR dip to 1.0\%/14.0\%. This mild degradation suggests that pushing the horizon too far begins to accumulate benign generation noise, which subsequently dilutes the malicious trigger's initial activation signal. Ultimately, this pattern indicates that the most discriminative evidence is concentrated in the first few decoding steps, making a short horizon both sufficient and computationally efficient.

\noindent \textbf{Effect of each module:}
To systematically evaluate the individual contributions of our proposed components in defending against diverse multimodal backdoor attacks, we conducted a comprehensive ablation study on the representative Qwen3-VL backbone. Results are shown in Table~\ref{tab:ablation}. The analysis across three representative attacks reveals three key insights: (1) relying on a static, single-step inspection severely cripples detection, as evidenced by the static Attention Flow (AF) variant failing almost completely, which proves that continuous early decoding trajectories capture far more robust backdoor evidence than isolated frozen snapshots; (2) neither AF nor Logits Flow (LF) alone is sufficient for comprehensive defense, whereas fusing these branches yields a substantial performance leap, confirming that attention patterns and logit distributions provide highly complementary traces of malicious behavior; and (3) converting this fused evidence into a reliable detector fundamentally relies on the Control-Aware Reference Normalization module. Without this normalization, the average F1 score plummets from 88.9\% to 68.2\%, and the model struggles most against the stealthy dual-trigger attack (dropping from 83.3\% to a mere 54.6\%). Overall, these results validate that our framework's robust performance stems directly from the synergy of dynamic trajectory profiling, multi-level feature fusion, and reference-normalized scoring.

\section{Conclusion}

In this paper, we rethink MLLM backdoor defense by shifting from static snapshots to decoding evolution. We introduce EVO-Detect, which exposes persistent hijacking mechanisms by jointly monitoring internal Attention Flow and interventional Logits Flow. To handle natural data complexity, a Control-Aware Reference Normalization mechanism anchors diagnostic signals against benign references, effectively suppressing false alarms. Extensive evaluations demonstrate that our modality-agnostic approach substantially outperforms previous attention-based baselines across diverse multimodal threats. This transition to temporal trajectory analysis provides a robust foundation for securing multimodal foundation models.

\begin{acks}
This work was supported by the National Natural Science Foundation of China under Grant 62576020.
\end{acks}

\clearpage
\bibliographystyle{ACM-Reference-Format}
\balance
\bibliography{mm2026_refs}

\end{document}